# A FUZZY MODEL FOR ANALOGICAL PROBLEM SOLVING

Michael Gr. Voskoglou

School of Technological Applications
Graduate Technological Educational Institute, Patras, Greece

`mvosk@hol.gr , voskoglou@teipat.gr`

## ABSTRACT

*In this paper we develop a fuzzy model for the description of the process of Analogical Reasoning by representing its main steps as fuzzy subsets of a set of linguistic labels characterizing the individuals' performance in each step and we use the Shannon- Wiener diversity index as a measure of the individuals' abilities in analogical problem solving. This model is compared with a stochastic model presented in author's earlier papers by introducing a finite Markov chain on the steps of the process of Analogical Reasoning. A classroom experiment is also presented to illustrate the use of our results in practice.*

## KEYWORDS

*Fuzzy Sets, Analogical Reasoning, Problem Solving*

## 1. INTRODUCTION

*Analogical Reasoning* (AR) is a method of processing information that compares the similarities between new and past understood concepts, then using these similarities to gain understanding of the new concept. The basic intuition behind AR is that when there are substantial parallels across different situations there are likely to be further parallels. AR is ubiquitous in human cognition. Analogies are used in explaining concepts which cannot directly perceived (e.g. electricity in terms of the water flow), in making predictions within domains, in communication and persuasion, etc. Within cognitive science mental processes are likened to computer programs (e.g. neural networks) and such analogies serve as mental models to support reasoning in new domains. AR is important in general in creativity and scientific discovery. In Artificial Intelligence the *Case-Based- Reasoning* paradigm covers a range of different methods (including analogical reasoning) for organizing, retrieving, utilizing and indexing the knowledge of past cases [15].

Solution of problems by analogy (*analogical problem solving*) is a special case of the general class of AR. However this strategy can be difficult to implement in problem solving, because it requires the solver to attend to information other than the problem to be solved (*target problem*). Thus the solver may come up empty-handed, either because he has not solved any similar problems in past, or because he fails to realize the relevance of previous problems. But, even if an analogue is retrieved, the solver must know how to use it to determine the solution procedure for the target problem.

Several studies ([1], [2], [3], [6], [10], etc) have provided detailed models for the process of AR which are broadly consistent with reviews of problem solving strategy training studies, in which factors associated with instances of successful transfer – that is, use of already existing knowledge to produce new knowledge - are identified. According to these studies the main steps (sub processes) involved in AR include:

- *Representation* of the target problem.
- *Search-retrieval* of a related past problem.
- *Mapping* of the representations of the target and the related problem.
- *Adaptation* of the solution of the related problem for use with the target problem.

More specifically, before solvers working on a problem they usually construct a representation of it. A good representation must include both the surface and the structural (abstract, solution relevant) features of the problem. The former are mainly determined by what are the quantities involved in the problem and the latter by how these quantities are related to each other. The features included in solvers' representations of the target problem are used as retrieval cues for a related problem in memory (called a *source problem*). When the two problems share structural but not surface structures the source is called a *remote analogue* of the target problem Analogical mapping requires *aligning* the two situations – that is, finding the correspondences between the representations of the target and the source problem – and *projecting inferences* from the source to the target. Once the common alignment and the candidate inferences have been discovered the analogy is *evaluated*. The last step involves the adaptation of the solution of the analogous problem for use with the target problem, where the correspondences between objects and relations of the two problems must be used.

The successful completion of the above process is referred as *positive analogical transfer*. But the search may also yield *distractor problems* having surface but not structural (solution relevant) common features with the target problem and therefore being only superficially similar to it. Usually the reason for this is a non satisfactory representation of the target problem, containing only its salient surface features, and the resulting consequences on the retrieval cues available for the search process. When a distractor problem is considered as an analogue of the target, we speak about *negative analogical transfer*. This happens if a distractor problem is retrieved as a source problem and the solver fails, through the mapping of the representations of the source and target problem, to realize that the source cannot be considered as an analogue to the target. Therefore the process of mapping is very important in analogical problem solving playing the role of a "*control system*" for the fitness of the source problem.

## 2. THE MODEL

Mathematics does not explain the natural behaviour of an object, it simply describes it. This description however is so much effective, so that an elementary mathematical equation can describe simply and clearly a relation that in order to be expressed with words could need entire pages. During the last two decades (1991-2011) we have worked in creating mathematical (stochastic and fuzzy) models for the better description and understanding of several processes appearing in the areas of Mathematical Education, Artificial Intelligence (Case-Based Reasoning) and Management [14]. Continuing this research in this paper we shall develop a fuzzy model for the description of the process of AR by representing its main steps as fuzzy subsets of a set of linguistic labels characterizing the individuals' performance in each of these steps. For general facts on fuzzy sets we refer freely to [4]. In order to make the development of our model technically simpler we shall consider the step of representation of the target problem as a sub step of the search-retrieval of the source problem.

Let us consider a group of *n* analogical problem solvers, $n \geq 2$. Denote by $A_i$, i=1,2,3 the steps of search-retrieval, mapping and adaptation respectively. Denote also by *a, b, c, d,* and *e* the linguistic labels of negligible, low, intermediate, high and complete success respectively of the analogical problem solvers in each of the $A_i$'s.

Set *U = {a, b, c, d, e}* and let $n_{ia}$, $n_{ib}$, $n_{ic}$, $n_{id}$ and $n_{ie}$ denote the number analogical problem solvers who faced negligible, low, intermediate, high and complete success at step $A_i$, i=1,2,3. We define the *membership function* $m_{Ai}$ for each x in U, as follows:

$$m_{A_i}(x) = \begin{cases} 1 & \text{if } \frac{4n}{5} < n_{ix} \leq n \\ 0{,}75 & \text{if } \frac{3n}{5} < n_{ix} \leq \frac{4n}{5} \\ 0{,}5 & \text{if } \frac{2n}{5} < n_{ix} \leq \frac{3n}{5} \\ 0{,}25 & \text{if } \frac{n}{5} < n_{ix} \leq \frac{2n}{5} \\ 0 & \text{if } 0 \leq n_{ix} \leq \frac{n}{5} \end{cases}$$

Then $A_i$ is represented as a fuzzy subset of U by $A_i = \{(x, m_{Ai}(x)): x \in U\}$, i=1, 2, 3.
In order to represent all possible solvers' *profiles (overall states)* during the AR process we consider a *fuzzy relation*, say R, in $U^3$ of the form

$$R = \{(s, m_R(s)): s=(x, y, z) \in U^3\}.$$

Since the degree of solvers' success at a certain step depends upon the degree of their success in the previous step and in order to determine properly the membership function $m_R$ we give the following definition:

A profile s=(x, y, z), with x, y, z in U, is said to be *well ordered* if x corresponds to a degree of success equal or greater than y, and y corresponds to a degree of success equal or greater than z. For example, *(c, c, a)* is a well ordered profile, while *(b, a, c)* is not. We define now the *membership degree* of a profile *s* to be

$m_R(s) = m_{A_1}(x) m_{A_2}(y) m_{A_3}(z)$, if s is well ordered, and *0* otherwise. In fact, if for example profile *(b, a, c)* possessed a nonzero membership degree, how it could be possible for a solver, who failed at the step of mapping, to perform satisfactorily at the step of adaptation?
Next, for reasons of brevity, we shall write $m_s$ instead of $m_R(s)$. Then the *possibility* $r_s$ of the profile s is defined by $r_s = \frac{m_s}{\max\{m_s\}}$, where *max {$m_s$}* denotes the maximal value of $m_s$, for all *s* in $U^3$. In other words the possibility of *s* expresses the "relative membership degree" of *s* with respect to *max {$m_s$}*. Calculating the possibilities of all profiles one obtains a qualitative view of the group's performance during the AR process.

Further, the amount of information obtained by an action can be measured by the reduction of uncertainty resulting from this action. Accordingly the indviduals' uncertainty during the AR process is connected to their capacity in obtaining relevant information. Therefore a measure of uncertainty could be adopted as a measure of the group's abilities in AR. For example, in earlier papers ([11], [12], [13]) developing analogous fuzzy models for learning, case-based reasoning and mathematical modelling we have used the *total possibilistic uncertainty* T as a measure of the students' abilities.

Another measure of (probabilistic) uncertainty and the associated information was established by Shannon [7]. When expressed in terms of the Dempster-Shafer mathematical theory of evidence, this measure takes the form H= $-\frac{1}{\ln n}\sum_{s=1}^{n} m_s \ln m_s$ , where n is the total number of elements of the corresponding fuzzy set ([5]; p.20), and it is called the *Shannon entropy* or the *Shannon- Wiener diversity index*. In the above formula the sum is divided by *ln n* in order to normalize H, so that its maximal value is 1 regardless the value of n.

Adopting H as a measure of the group's abilities in AR it becomes evident that the lower is the value of H (i.e. the higher is the reduction of the corresponding uncertainty), the better the group's abilities. An advantage of adopting H as a measure instead of T is that H is calculated directly from the membership degrees of all profiles *s*, in contrast to T that presupposes the calculation of the possibilities of all profiles first.

Assume now that one wants to study the *combined results of behaviour* of *k* different groups, *k≥2*, during the same process. For this we introduce the *fuzzy variables $A_1(t)$, $A_2(t)$ and $A_3(t)$* with *t=1, 2,…, k*. The values of these variables represent fuzzy subsets of *U* corresponding to the steps of the AR process for each of the k groups; e.g. $A_1(2)$ represents the fuzzy subset of *U* corresponding to the step of search-retrieval for the second group (t=2). It becomes evident that, in order to measure the degree of evidence of combined results of the *k* groups, it is necessary to define the possibility *r(s)* of each profile *s* with respect to its membership degrees for all groups. For this, we introduce the *pseudo-frequencies* f(s) = $\sum_{t=1}^{k} m_s(t)$ and we define r(s) = $\frac{f(s)}{\max\{f(s)\}}$, where *max {f(s)}* denotes the maximal pseudo-frequency. Obviously the same method could be applied when one wants to study the combined results of behaviour of a group during different analogical problem solving processes.

## 3. AN APPLICATION IN THE CLASSROOM

In order to illustrate the use of the above model in practice we performed the following experiment, where the subjects were students of the graduate Technological Educational Institute of Patras, Greece, being at their second term of studies. We formed two groups, with 20 students of the School of Management and Economics in the first and 20 students of the School of Technological Applications (prospective engineers) in the second group.

Four mathematical problems were given for solution to both groups (see Appendix). The first of these problems was related to combinatorial analysis and probability, the second was related to the theory of matrices and the third one was an application of derivatives to economics. All the above are topics in the students' first term course of mathematics. The fourth was a problem involving arithmetic i.e., an already developed at school ability. In each case and before receiving the target problem students received three other problems together with their solution procedures. They read each problem and its solution procedure and then solved the problem themselves using the given procedure. Subjects were allowed 10 minutes for each problem and they were not given the other problem until after 10 minutes had elapsed. The first of these problems was a remote analogue to the target problem, the second was a distractor problem, while the third was unrelated to the target problem in terms of both their surface and structure features. Next the target problem was given and was asked from the subjects to try to solve it by adapting the solution of one of the previous problems (time allowed 20 minutes). Our instructions stressed the importance of showing all of one's work on paper and emphasized that we were interested in both correct and incorrect solution attempts.

Our characterizations of students' performance at <u>each step</u> of the AR process involved:

- Negligible success, if they didn't obtain (at the particular step) positive results for the given problems.
- Low success, if they obtained positive results for 1 only of the given problems.
- Intermediate success, if they obtained positive results for 2 problems.
- High success, if they obtained positive results for 3 problems.
- Complete success, if they obtained positive results for all the given (4 in total) problems.

Examining students' papers of the first group we found that 9, 6 and 5 students had intermediate, high and complete success respectively at the step of search-retrieval in terms of choosing the correct problem (i.e. the remote analogue to the target) as the source problem. This means that $n_{1a}=n_{1b}=0$, $n_{1c}=9$, $n_{1d}=6$ and $n_{1e}=5$. Thus, according to the definition of $m_{A_i}(x)$, the step of search-retrieval corresponds to a fuzzy subset of U of the form: $A_1 = \{(a,0),(b,0),(c, 0.5),(d, 0.25),(e, 0.25)\}$.

In the same way we represented the steps of mapping and adaptation as fuzzy subsets
of U by $A_2 = \{(a,0),(b,0),(c, 0.5),(d, 0.25),(e,0)\}$ and
$A_3 = \{(a, 0.25),(b, 0.25),(c, 0.25),(d,0),(e,0)\}$ respectively.
Next, we calculated the membership degrees of the $5^3$ (ordered samples with replacement of 3 objects taken from 5) in total possible students' profiles (see column of $m_s(1)$ in Table 1). For example, for $s=(c, c, a)$ one finds that
$m_s = m_{A_1}(c) \cdot m_{A_2}(c) \cdot m_{A_3}(a) = (0.5).(0.5).(0.25) = 0.06225$.
It turned out that $(c, c, a)$ was one of the profiles possessing the maximal membership degree and therefore the possibility of each s in $U^3$ is given by $r_s = \dfrac{m_s}{0{,}06225}$. Using this formula we calculated the possibilities of all profiles (see column of $r_s(1)$ in Table 1).

Table 1. Profiles with non zero pseudo-frequencies

| $A_1$ | $A_2$ | $A_3$ | $m_s(1)$ | $r_s(1)$ | $m_s(2)$ | $r_s(2)$ | $f(s)$ | $r(s)$ |
|---|---|---|---|---|---|---|---|---|
| b | b | b | 0 | 0 | 0.016 | 0.258 | 0.016 | 0.129 |
| b | b | a | 0 | 0 | 0.016 | 0.258 | 0.016 | 0.129 |
| b | a | a | 0 | 0 | 0.016 | 0.258 | 0.016 | 0.129 |
| c | c | c | 0.062 | 1 | 0.062 | 1 | 0.124 | 1 |
| c | c | a | 0.062 | 1 | 0.062 | 1 | 0.124 | 1 |
| c | c | b | 0 | 0 | 0.031 | 0.5 | 0.031 | 0.25 |
| c | a | a | 0 | 0 | 0,031 | 0.5 | 0.031 | 0.25 |
| c | b | a | 0 | 0 | 0.031 | 0.5 | 0.031 | 0.25 |
| c | b | b | 0 | 0 | 0.031 | 0.5 | 0.031 | 0.25 |
| d | d | a | 0.016 | 0.258 | 0 | 0 | 0.016 | 0.129 |
| d | d | b | 0.016 | 0.258 | 0 | 0 | 0.016 | 0.129 |
| d | d | c | 0.016 | 0.258 | 0 | 0 | 0.016 | 0.129 |
| d | a | a | 0 | 0 | 0.016 | 0.258 | 0.016 | 0.129 |
| d | b | a | 0 | 0 | 0.016 | 0.258 | 0.016 | 0.129 |
| d | b | b | 0 | 0 | 0.016 | 0.258 | 0.016 | 0.129 |
| d | c | a | 0.031 | 0.5 | 0.031 | 0.5 | 0.062 | 0.5 |
| d | c | b | 0.031 | 0.5 | 0.031 | 0.5 | 0.062 | 0.5 |
| d | c | c | 0.031 | 0.5 | 0.031 | 0.5 | 0.062 | 0.5 |
| e | c | a | 0.031 | 0.5 | 0 | 0 | 0.031 | 0.25 |
| e | c | b | 0.031 | 0.5 | 0 | 0 | 0.031 | 0.25 |
| e | c | c | 0.031 | 0.5 | 0 | 0 | 0.031 | 0.25 |
| e | d | a | 0.016 | 0.258 | 0 | 0 | 0.016 | 0.129 |
| e | d | b | 0.016 | 0.258 | 0 | 0 | 0.016 | 0.129 |
| e | d | c | 0.016 | 0.258 | 0 | 0 | 0.016 | 0.129 |

(The outcomes of table 1 are written with accuracy up to the third decimal point)

Finally we calculated the Shannon entropy in terms of the values of column $m_s(1)$ in Table 1, where n=125 and we found that H=0,289.

Working as above for the second group we found that

$A_1=\{(a,0),(b, 0.25),(c, 0.5),(d, 0.25),(e,0)\}$,
$A_2=\{(a, 0.25),(b, 0.25),(c, 0.5),(d, 0),(e,0)\}$ and
$A_3=\{(a, 0.25),(b, 0.25),(c,0.25),(d,0),(e,0)\}$.

The membership degrees of all possible profiles of the second group are shown in column of $m_s(2)$ of Table 1. It turned out that the maximal membership degree was again 0.06225, therefore the possibility of each s is calculated by the same formula as for the first group. The possibilities of all profiles are shown in column of $r_s(2)$ of Table 1, while for the Shannon entropy we found that H=0,312. Thus, since *0,289<0,312*, the general performance of the first group was slightly better.

Next, in order to study the combined results of behaviour of the two groups, we introduced the fuzzy variables $A_i(t)$, i=1, 2, 3 and t=1, 2, as we have described them in the model. Then the pseudo-frequency of each student profile s is given by $f(s) = m_s(1) + m_s(2)$ (see corresponding column in Table 1). It turns out that the highest pseudo-frequency is 0,124 and therefore the possibility of each student's profile is given by $r(s) = \dfrac{f(s)}{0,124}$. The possibilities of all profiles having non-zero pseudo-frequencies are presented in the last column of Table 1.

## 4. DISCUSSION AND CONCLUSIONS

In this paper we presented a fuzzy model for the description of the process of AR by representing its main steps as fuzzy subsets of a set of linguistic labels characterizing the individuals' performance in each step and we used the Shannon- Wiener diversity index as a measure of the individuals' abilities in analogical problem solving. In earlier papers ([8], [9]) we have also developed a stochastic model for the same purposes by introducing a finite Markov chain on the steps of the process of AR. Both models give important quantitative information about the abilities of a group of analogical problem solvers'. The individuals have usually to search in their memories to retrieve the source among several past problems sharing common surface and/or structural characteristics with the target. However, in the stochastic model the more are the problems among which the source problem must be retrieved the more complicated becomes the calculation of the transition probabilities between states of the chain, because the individuals' "movements" in this case are extended to more directions. On the contrary, there is not any particular difficulty in this case with the fuzzy model. Moreover the fuzzy model gives a qualitative view of the group's performance through the calculation of the possibilities of all individuals' profiles during the AR process. Finally, an additional advantage of the fuzzy model is that it gives to the researcher the opportunity to study the combined results of the behaviour of two or more groups during the AR process or alternatively to study the combined results of the behaviour of the same group during different analogical problem solving processes. On the other hand the characterization of the analogical problem solvers' performance in terms of a set of linguistic labels which are fuzzy themselves is a disadvantage of the fuzzy model, because this characterization depends on the researcher's personal criteria (see for example in section 3 the criteria used in our experiments). Therefore a combined use of the two models seems to be the best solution in achieving a worthy of credit mathematical analysis of the AR process.

## APPENDIX

**Problems used in the classroom experiment**

CASE 1
Target problem: A box contains 8 balls numbered from 1 to 8. One makes three successive drawings, putting back the corresponding ball to the box before the next drawing. Find the probability of getting all the balls drawing out of the box different to each other. –

The probability is equal to the quotient of the total number of the ordered samples of 3 objects from 8 (favourable outcomes) to the total number of the corresponding samples with replacement (possible outcomes).
Remote analogue: How many numbers of 2 digits can be formed by using the digits from 1 to 6 and how many of them have their digits different?

Solution procedure given to the students: Find the total number of the ordered samples of 2 objects from 6 with and without replacement respectively.

Distractor problem: A box contains 3 white, 4 blue and 6 black balls. If we draw out 2 balls, what is the probability to be of the same colour?

Solution procedure given to the students: The number of all favourable outcomes is equal to the sum of the total number of combinations of 3, 4 and 6 objects taken 2 at each time respectively, while the number of all possible outcomes is equal to the total number of combinations of 13 objects taken 2 at each time.

Unrelated problem: Find the number of all possible anagrammatisms of the word "SIMPLE". How many of them start with S and how many of them start with S and end with E?

Solution procedure given to the students: The number of all possible anagrammatisms is equal to the total number 6! of permutations of 6 objects. The anagrammatisms starting with S are 5! And the anagrammatisms starting with S and ending with E are 4!

CASE 2
Target problem: Consider the matrices:

$$A = \begin{bmatrix} 1 & -α & -α \\ 0 & 1 & -α \\ 0 & 0 & 1 \end{bmatrix} \text{ και } B = \begin{bmatrix} 0 & -α & -α \\ 0 & 0 & -α \\ 0 & 0 & 0 \end{bmatrix}.$$

Prove that $A^n = A + (n-1)(B + \frac{n}{2} B)$, for every positive integer $n$.-

Since A=I+B, where I stands for the unitary 3X3 matrix, and $B^3 = 0$, is $A^n = (I+B)^n = I + nB + \frac{n(n-1)}{2} B^2 == A + (n-1)B + \frac{n(n-1)}{2} B = A + (n-1)(B + \frac{n}{2} B)$ .

Remote analogue : Let α be a nonzero real number. Prove that $α^n = \sum_{i=0}^{n} \binom{n}{i} (a-1)^n$, for all positive integers $n$.

Solution procedure given to the students: Write α = 1+(α-1) and apply the Newton's formula $(x+b)^n = \sum_{i=0}^{n} \binom{n}{i} x^{n-i} b^i$, setting x=1 and b=α-1.

Distractor problem : If A and B are as in the target problem, calculate $(A+B)^2$.
The students were asked to operate the corresponding calculations..

Unrelated problem: Prove that $1+2+\ldots+n=\dfrac{n(n+1)}{2}$, for all positive integers *n*.
The students were asked to apply induction on *n*.

CASE 3
Target problem: The price of sale of a good depends upon its total demand Q and it is given by $P(Q) = \frac{1}{2}Q-50$, while the cost of production of the good is given by
$C(Q)= \frac{1}{4}Q^2 +35Q+25$. Find the quantity Q of the good's total demand maximizing the profit from sale.-

The revenue from sale is equal to P(Q)Q and therefore the profit from sale is given by K(Q) = P(Q)Q-C(Q). The maximum of function K(Q) is calculated by using the well known theorem of derivatives.

Remote analogue: A car is entering to a road having initial speed 50 Km/h, which is changed according to the relation $U(t)=3t^2-12t+50$, where t represents the time (in minutes) during which the car is moving on this road. Find the minimal speed of the car on this road.-
The students were asked to apply the well known theorem of derivatives in order to calculate the minimum of the function U(t).

Distractor problem: The price of sale of a good depends upon its total demand Q and it is given by $P(Q)=25-Q^2$. The price is finally fixed to 9 monetary units and therefore the consumers who would be willing to pay more than this price benefit. Find the total benefit to consumers (Dowling 1980, paragraph 17.7: *Consumer's surplus*).

Solution procedure given to the students: For P=9 and since $Q \geq 0$, it turns out that Q=4.metric units. Drawing the graph of the function P(Q) (parabola) it is easy to observe that the total benefit to consumers is equal to $\int_0^4 P(Q)dQ$ - 4.9 monetary units

Unrelated problem: Find the area under the curve $y=4x^2+2$.

Solution procedure given to the students: The area is given by $\int (4x^2 + 2)dx$.

CASE 4
Target problem: A producer has a stock of wine greater than 500 and less than 750 kilos. He has estimated that, if he had the double quantity of wine and transfused it to bottles of 12, or 25, or 40 kilos, it would be left over 6 kilos at each time. Find the quantity of the stock.-
If Q is the quantity of stock, then, since the lowest common multiple of 12,25 and 40 is 600, 2Q-6 is a multiple of 600, therefore 2Q=606, or 2Q=1212, or 2Q=1818, etc. But 500<Q<750, therefore Q=603 kilos.

Remote analogue: An employer occupies less than 50 workers. If he occupied the triple number of workers and 3 more, then he could distribute them in bands of 8 or 12 or 15 workers. How many workers he occupies?

Solution procedure given to the students: If x is the number of workers, then , since the lowest common multiple of 8, 12 and 15 is 120, 3x+3 is a multiple of 120, or x+1 is a multiple of 40.
Distractor problem: A producer has a stock of 3400 and 5025 kilos respectively of two different kinds of wine and he decides to distribute these quantities to the maximal possible number of customers. After this distribution, they remained 25 kilos from each kind of wine in his barrels. How many of his customers he succeeded to satisfy with this manipulation?

Solution procedure given to the students: The number of customers is equal to the greatest common divisor of 3400-25 and 5025-25.

Unrelated problem: The number of students of a school is between 300 and 400. When they tried marching in rows of 10 the last row had 9 students, while when they tried marching in rows of 9 the last row had 7 students. How many are the students?

Solution procedure given to the students: Let a=100x+10y+z be the number of students of the school. Then a=10t-1 for some positive integer t. Therefore z=9 and x=3. Further a=9s+7 for some positive integer s, or a-7=9s. But, since 9 divides a-7, 9 divides also the sum of the digits of a-7, i.e. (3+y+9)-7=9k for some positive integer k, or y+5=9k. But 0<y≤9, therefore y=4.

## Author

**Michael Gr. Voskoglou** (B.Sc., M.Sc., M.Phil. , Ph.D. in Mathematics) is currently Professor of Mathematical Sciences at the Graduate Technological Educational Institute of Patras, Greece. He is the author of 8 books (7 in Greek and 1 in English language) and of more than 200 papers published in reputed journals of 22 countries, with many references from other researchers. He is also a reviewer of the AMS and member of the Editorial Board or referee in several mathematical journals.

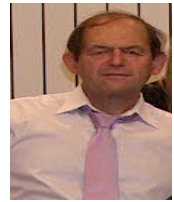